\pdfoutput=1

\documentclass{article}
\usepackage{ijcai17}

\usepackage{times}

\usepackage{latexsym} 
\usepackage{enumerate}
\usepackage{bm}
\usepackage{algorithm}
\usepackage{algorithmicx}
\usepackage{algpseudocode}
\usepackage{dsfont}
\usepackage{bbm}
\usepackage{verbatim}
\usepackage{arydshln}
\usepackage{multirow}
\usepackage{multicol}
\usepackage{amsmath}
\usepackage{paralist}
\usepackage{color}
\usepackage{graphicx}
\usepackage{subfigure}
\usepackage{hyperref}

\newcommand{\ignore}[1]{{}}

\def\ours{{\tt LTSG}}
\def\lda{{\tt LDA}}
\def\mlda{{\tt MRF-LDA}}
\def\twe{{\tt TWE}}

\title{LTSG: Latent Topical Skip-Gram for Mutually Learning Topic Model and Vector Representations}

\author{Jarvan Law, Hankz Hankui Zhuo, Junhua He and Erhu Rong \\
	Dept. of Computer Science, Sun Yat-Sen University, GuangZhou, China. 510006 \\
 	{\tt JarvanLaw@gmail.com}, {\tt zhuohank@mail.sysu.edu.cn} \\
 	{\tt \{hejunh,rongerhu\}@mail2.sysu.edu.cn}  
  }

\begin{document}

\maketitle

\begin{abstract}
Topic models have been widely used in discovering latent topics which are shared across documents in text mining. Vector representations, word embeddings and topic embeddings, map words and topics into a low-dimensional and dense real-value vector space, which have obtained high performance in NLP tasks. However, most of the existing models assume the result trained by one of them are perfect correct and used as prior knowledge for improving the other model. Some other models use the information trained from external large corpus to help improving smaller corpus. In this paper, we aim to build such an algorithm framework that makes topic models and vector representations mutually improve each other within the same corpus. An EM-style algorithm framework is employed to iteratively optimize both topic model and vector representations. Experimental results show that our model outperforms state-of-art methods on various NLP tasks.
\end{abstract}

\section{Introduction}
\label{intro}
Word embeddings, e.g., distributed word representations \cite{DBLP:conf/nips/MikolovSCCD13}, represent words with low dimensional and dense real-value vectors, which capture useful semantic and syntactic features of words. Distributed word embeddings can be used to measure word similarities by computing distances between vectors, which have been widely used in various IR and NLP tasks, such as entity recognition \cite{DBLP:conf/acl/TurianRB10}, disambiguation \cite{DBLP:journals/jmlr/CollobertWBKKK11} and parsing \cite{DBLP:conf/icml/SocherLNM11,DBLP:conf/acl/SocherBMN13}. Despite the success of previous approaches on word embeddings, they all assume each word has a specific meaning and represent each word with a single vector, which restricts their applications in fields with polysemous words, e.g., ``bank'' can be either ``a financial institution'' or ``a raised area of ground along a river''.

To overcome this limitation, \cite{DBLP:conf/aaai/LiuLCS15} propose a topic embedding approach, namely Topical Word Embeddings ({\twe}), to learn topic embeddings to characterize various meanings of polysemous words by concatenating topic embeddings with word embeddings. Despite the success of {\twe}, compared to previous multi-prototype models \cite{DBLP:conf/naacl/ReisingerM10,DBLP:conf/acl/HuangSMN12}, it assumes that word distributions over topics are provided by off-the-shelf topic models such as {\lda}, which would limit the applications of {\twe} once topic models do not perform well in some domains \cite{DBLP:conf/nips/PettersonSCBN10,DBLP:journals/tkde/PhanNLNHH11}. As a matter of fact, pervasive polysemous words in documents would harm the performance of topic models that are based on co-occurrence of words in documents. Thus, a more realistic solution is to build both topic models with regard to polysemous words and polysemous word embeddings simultaneously, instead of using off-the-shelf topic models.

\begin{figure}
\label{model-framework}
\includegraphics[width=0.48\textwidth]{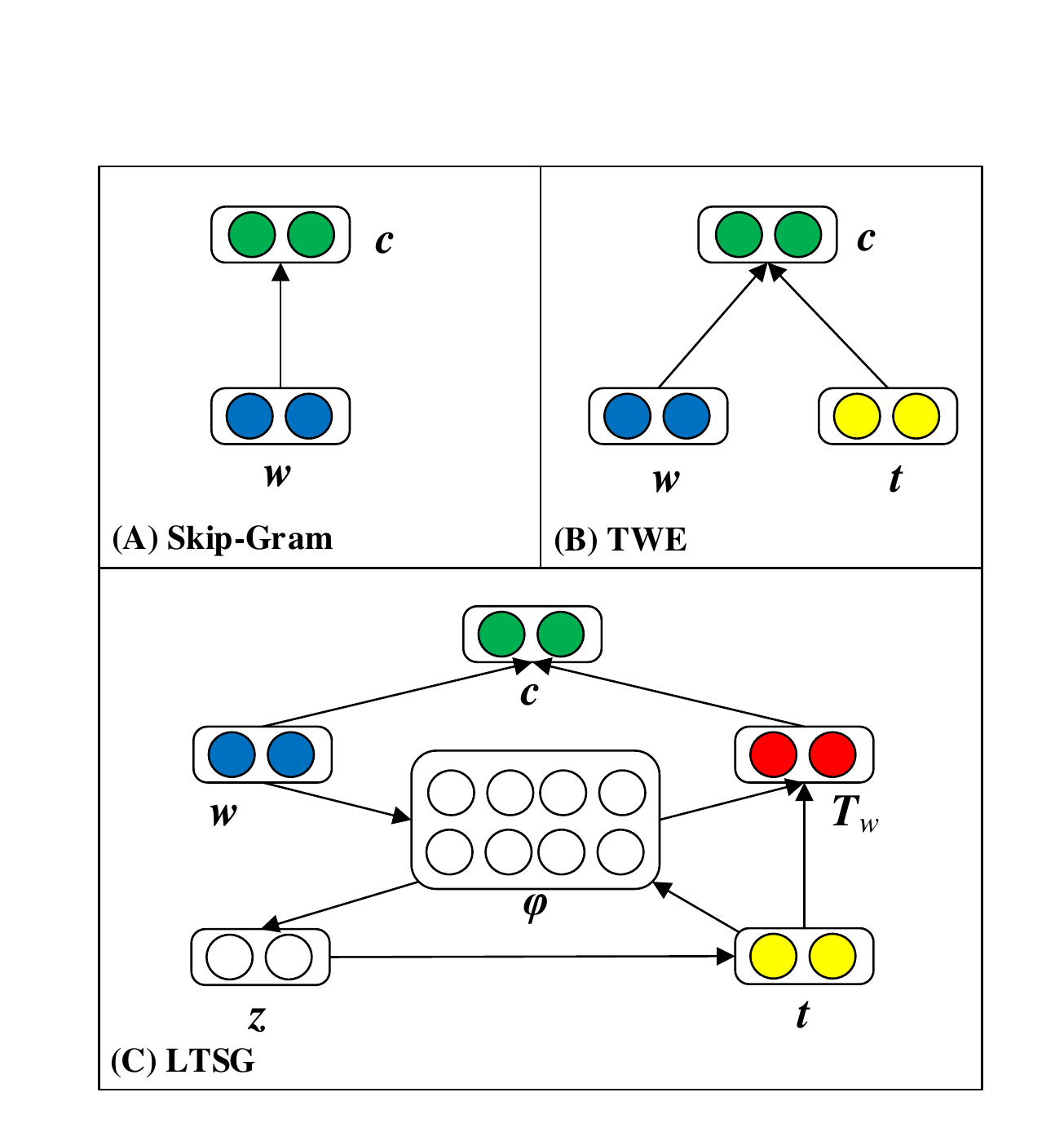}
\caption{Skip-Gram, {\twe} and {\ours} models. Blue, yellow, green circles denote the embeddings of word, topic and context, while red circles in {\ours} denote the global topical word. White circles denote the topic model part, topic-word distribution $\bm{\varphi}$ and topic assignment $\bm{z}$.}
\end{figure}

In this work, we propose a novel learning framework, called Latent Topical Skip-Gram ({\ours}) model, to mutually learn polysemous-word models and topic models. To the best of our knowledge, this is the first work that considers learning polysemous-word models and topic models simultaneously. Although there have been approaches that aim to improve topic models based on word embeddings {\mlda} \cite{DBLP:conf/naacl/XieYX15}, they fail to improve word embeddings provided words are polysemous; although there have been approaches that aim to improve polysemous-word models {\twe} \cite{DBLP:conf/aaai/LiuLCS15} based on topic models, they fail to improve topic models considering words are polysemous. Different from previous approaches, we introduce a new node $\bm{T_}w$, called \emph{global topic}, to capture all of the topics regarding polysemous word $w$ based on topic-word distribution $\bm{\varphi}$, and use the global topic to estimate the context of polysemous word $w$. Then we characterize polysemous word embeddings by concatenating word embeddings with topic embeddings. We illustrate our new model in Figure \ref{model-framework}, where Figure \ref{model-framework}(A) is the skip-gram model \cite{DBLP:conf/nips/MikolovSCCD13}, which aims to maximize the probability of context $c$ given word $w$, Figure \ref{model-framework}(B) is the {\twe} model, which extends the skip-gram model to maximize the probability of context $c$ given both word $w$ and topic $t$, and Figure \ref{model-framework}(C) is our {\ours} model which aims to maximize the probability of context $c$ given word $w$ and global topic $\bm{T_}w$. $\bm{T_}w$ is generated based on topic-word distribution $\bm{\varphi}$ (i.e., the joint distribution of topic embedding $\bm{t}$ and word embedding $\bm{w}$) and topic embedding $\bm{t}$ (which is based on topic assignment $\bm{z}$). Through our {\ours} model, we can simultaneously learn word embeddings $\bm{w}$ and global topic embeddings $\bm{T_}w$ for representing polysemous word embeddings, and topic word distribution $\bm{\varphi}$ for mining topics with regard to polysemous words. We will exhibit the effectiveness of our {\ours} model in text classification and topic mining tasks with regard to polysemous words in documents.

In the remainder of the paper, we first introduce preliminaries of our {\ours} model, and then present our {\ours} algorithm in detail. After that, we evaluate our {\ours} model by comparing our {\ours} algorithm to state-of-the-art models in various datasets. Finally we review previous work related to our {\ours} approach and conclude the paper with future work.

\section{Preliminaries}
In this section, we briefly review preliminaries of Latent Dirichlet Allocation ({\lda}), Skip-Gram, and Topical Word Embeddings ({\twe}), respectively. We show some notations and their corresponding meanings in Table \ref{terminologies}, which will be used in describing the details of {\lda}, Skip-Gram, and {\twe}.
\begin{table*}[!ht]
\caption{Notations of the text collection.}\label{terminologies} 
\begin{center}
\begin{tabular}{|c|c|l|}
\hline \textbf{Term} & \textbf{Notation} & \multicolumn{1}{c|}{\textbf{Definition or Description}} \\ \hline
\emph{vocabulary} & $\bm{\mathcal{V}}$ & set of words in the text collection, $\vert\bm{\mathcal{V}}\vert = W$\\
\emph{word} & $w$ & a basic item from vocabulary indexed as $w \in \{1,2,\ldots,W\}$ \\
\emph{document} & $\mathbf{w}$ & a sequence of $N$ words, $\mathbf{w} = (w_1,w_2,\ldots,w_N)$ \\
\emph{corpus} & $\bm{\mathcal{D}}$ & a collection of $M$ documents, $\bm{\mathcal{D}} = \{\mathbf{w}_1, \mathbf{w}_2,\ldots,\mathbf{w}_M\}$ \\ 
\hline
\emph{topic-word} & $\bm{\varphi}$ & $K$ distributions over vocabulary ($K \times W$ matrix), $\vert\bm{\varphi}\vert=K, \vert\bm{\varphi}_k\vert = W$ \\ 
\emph{word embedding} & $\bm{v}$ & distributed representation of $word$, denoted by $\bm{v}_w$, $\bm{v} \in \mathbbm{R}^d$ \\
\emph{topic embedding} & $\bm{t}$ & distributed representation of $topic$, denoted by $\bm{t}_k$, $\bm{t} \in \mathbbm{R}^d$ \\
\hline
\end{tabular}
\end{center}
\end{table*}

\subsection{Latent Dirichlet Allocation}
\label{LDA}
Latent Dirichlet Allocation ({\lda}) \cite{DBLP:journals/jmlr/BleiNJ03}, a three-level hierarchical Bayesian model, is a well-developed and widely used probabilistic topic model. Extending Probabilistic Latent Semantic Indexing (PLSI) \cite{DBLP:conf/sigir/Hofmann99}, {\lda} adds Dirichlet priors to document-specific topic mixtures to overcome the overfitting problem in PLSI. {\lda} aims at modeling each document as a mixture over sets of topics, each associated with a multinomial word distribution. Given a document corpus $\bm{\mathcal{D}}$, each document $\mathbf{w}_m \in \bm{\mathcal{D}}$ is assumed to have a distribution over $K$ topics. The generative process of {\lda} is shown as follows,
\begin{enumerate}
\item For each topic $k = 1 \to K$,
	draw a distribution over words $\bm{\varphi}_k \sim Dir(\bm{\beta})$
\item For each document $\mathbf{w}_m \in \bm{\mathcal{D}}, m \in \{1,2,\ldots,M\}$
	\begin{enumerate}
	\item Draw a topic distribution $\bm{\theta}_m \sim Dir(\bm{\alpha})$
	\item For each word $w_{m,n} \in \mathbf{w}_m, n = 1,\ldots,N_m$
		\begin{enumerate}
		\item Draw a topic assignment $z_{m,n} \sim Mult(\bm{\theta}_m)$, $z_{m,n} \in \{1,\ldots, K\}.$
		\item Draw a word $w_{m,n} \sim Mult(\bm{\varphi}_{z_{m,n}})$
		\end{enumerate}
	\end{enumerate}
\end{enumerate}
where $\bm{\alpha}$ and $\bm{\beta}$ are Dirichlet hyperparameters, specifying the nature of priors on $\bm{\theta}$ and $\bm{\varphi}$. Variational inference and Gibbs sampling are the common ways to learn the parameters of {\lda}.

\subsection{The Skip-Gram Model}
\label{SG}
The Skip-Gram model is a well-known framework for learning word vectors \cite{DBLP:conf/nips/MikolovSCCD13}. Skip-Gram aims to predict context words given a target word in a sliding window, as shown in Figure \ref{model-framework}(A). 

Given a document corpus $\bm{\mathcal{D}}$ defined in Table \ref{terminologies}, the objective of Skip-Gram is to maximize the average log-probability
\begin{eqnarray}
\mathcal{L}(\bm{\mathcal{D}}) = \frac{1}{\sum_{m=1}^M N_m} \sum\limits_{m=1}^M \sum\limits_{n=1}^{N_m} \sum\limits_{-c\leq j\leq c, j\neq 0} \notag\\ \log\Pr(w_{m,n+j} \vert w_{m,n}),
\end{eqnarray}
where $c$ is the context window size of the target word. The basic Skip-Gram formulation defines $\Pr(w_{m,n+j} \vert w_{m,n})$ using the softmax function:
\begin{equation}
\Pr(w_{m,n+j} \vert w_{m,n}) = \frac{\exp (\bm{v}_{w_{m,n+j}} \cdot \bm{v}_{w_{m,n}})}{\sum_{w=1}^W \exp (\bm{v}_w \cdot \bm{v}_{w_{m,n}})},
\end{equation}
where $\bm{v}_{w_{m,n}}$ and $\bm{v}_{w_{m,n+j}}$ are the vector representations of target word $w_{m,n}$ and its context word $w_{m,n+j}$, and $W$ is the number of words in the vocabulary $\bm{\mathcal{V}}$. Hierarchical softmax and negative sampling are two efficient approximation methods used to learn Skip-Gram.

\subsection{Topical Word Embeddings}
\label{TWE}
Topical word embeddings ({\twe}) is a more flexible and powerful framework for multi-prototype word embeddings, where topical word refers to a word taking a specific topic as context \cite{DBLP:conf/aaai/LiuLCS15}, as shown in Figure \ref{model-framework}(B). {\twe} model employs {\lda} to obtain the topic distributions of document corpora and topic assginment for each word token. {\twe} model uses topic $z_{m,n}$ of target word to predict context word compared with only using the target word $w_{m,n}$ to predict context word in Skip-Gram. {\twe} is defined to maximize the following average log probability
\begin{equation}
\begin{split}
&\mathcal{L}(\bm{\mathcal{D}}) = \frac{1}{\sum_{m=1}^M N_m} \sum\limits_{m=1}^M \sum\limits_{n=1}^{N_m} \sum\limits_{-c\leq j\leq c, j\neq 0} \\ &\log\Pr(w_{m,n+j} \vert w_{m,n}) + \log\Pr(w_{m,n+j} \vert z_{m,n}).
\end{split}
\end{equation}
{\twe} regards each topic as a pseudo word that appears in all positions of words assigned with this topic. When training {\twe}, Skip-Gram is being used for learning word embeddings. Afterwards, each topic embedding is initialized with the average over all words assigned to this topic and learned by keeping word embeddings unchanged. 

Despite the improvement over Skip-Gram, the parameters of {\lda}, word embeddings and topic embeddings are learned separately. In other word, {\twe}  just uses {\lda} and Skip-Gram to obtain external knowledge for learning better topic embeddings.

\section{Our {\ours} Algorithm}
\label{LTSG}
Extending from the {\twe} model, the proposed Latent Topical Skip-Gram model (\ours) directly integrates {\lda} and Skip-Gram by using topic-word distribution $\bm{\varphi}$ mentioned in topic models like {\lda}, as shown in Figure \ref{model-framework}(C). We take three steps to learn topic modeling, word embeddings and topic embeddings simultaneously, as shown below.
\begin{enumerate}[\textbf{Step} 1]
\item \label{Step-1}
\textbf{Sample topic assignment for each word token.} Given a specific word token $w_{m,n}$, we sample its latent topic $z_{m,n}$ by performing Gibbs updating rule similar to {\lda}. 
\item \label{Step-2}
\textbf{Compute topic embeddings.} We average all words assigned to each topic to get the embedding of each topic.
\item \label{Step-3}
\textbf{Train word embeddings.} We train word embeddings similar to Skip-Gram and {\twe}. Meanwhile, topic-word distribution $\bm{\varphi}$ is updated based on Equation (\ref{phi_update}). The objective of this step is to maximize the following function
\begin{equation}
\label{LTSG_OBJFUNC}
\begin{split}
&\mathcal{L}(\bm{\mathcal{D}}) = \frac{1}{\sum_{m=1}^M N_m} \sum\limits_{m=1}^M \sum\limits_{n=1}^{N_m} \sum\limits_{-c\leq j\leq c, j\neq 0} \\ &\log\Pr(w_{m,n+j} \vert w_{m,n}) + \log\Pr(w_{m,n+j} \vert T_{w_{m,n}}),
\end{split}
\end{equation}
where $T_{w_{m,n}} = \sum\limits_{k=1}^K \bm{t}_k \cdot \varphi_{k,w_{m,n}}$. $\bm{t}_k$ indicates the $k$-th topic embedding. $T_{w_{m,n}}$ can be seen as a distributed representation of global topical word of $w_{m,n}$.
\end{enumerate}

We will address the above three steps in detail below. 

\subsection{Topic Assignment via Gibbs Sampling}
To perform Gibbs sampling, the main target is to sample topic assingments $z_{m,n}$ for each word token $w_{m,n}$. Given all topic assignments to all of the other words, the full conditional distribution $\Pr(z_{m,n} = k \vert \mathbf{z}^{-(m,n)}, \mathbf{w})$ is given below when applying collapsed Gibbs sampling \cite{griffiths2004finding},
\begin{equation}
\label{GibbsLDA}
\begin{split}
\Pr(z_{m,n} = k \vert \mathbf{z}^{-(m,n)}, \mathbf{w}) \propto & \frac{n_{k,w_{m,n}}^{-(m,n)} + \beta}{\sum_{w=1}^w n_{k,w}^{-(m,n)} + W\beta} \\
 &\cdot \frac{n_{m,k}^{-(m,n)} + \alpha}{\sum_{k^\prime=1}^K n_{m,k^\prime}^{-(m,n)} + K\alpha},
\end{split}
\end{equation}
where $-(m,n)$ indicates that the current assignment of $z_{m,n}$ is excluded. $n_{k,w}$ and $n_{m,k}$ denote the number of word tokens $w$ assigned to topic $k$ and the count of word tokens in document $m$ assinged to topic $k$, respectively. After sampling all the topic assignments for words in corpus $\bm{\mathcal{D}}$, we can estimate each component of $\bm{\varphi}$ and $\bm{\theta}$ by Equations (\ref{Phi_cal}) and (\ref{Theta_cal}).
\begin{equation}
\label{Phi_cal}
\hat{\varphi}_{k,w} = \frac{n_{k,w} + \beta}{\sum_{w^\prime=1}^W n_{k,w^\prime} + W\beta}
\end{equation}
\begin{equation}
\label{Theta_cal}
\hat{\theta}_{d,k} = \frac{n_{m,k} + \alpha}{\sum_{k^\prime=1}^K n_{m,k^\prime} + K\alpha}
\end{equation}

Unlike standard {\lda}, the topic-word distribution $\bm{\varphi}$ is used directly for constructing the modified Gibbs updating rule in {\ours}. Following the idea of DRS \cite{DBLP:conf/ijcai/DuJSL15}, with the conjugacy property of Dirichlet and multinomial distributions, the Gibbs updating rule of our model {\ours} can be approximately represented by 
\begin{equation}
\label{sampler}
\begin{split}
\Pr(z_{m,n} =& k \vert \mathbf{w, z}^{-(m,n)}, \bm{\varphi, \alpha}) \propto \\ 
& \varphi_{k,w_{m,n}} \cdot \frac{n_{m,k}^{-(m,n)} + \alpha}{\sum_{k^\prime=1}^K n_{m,k^\prime}^{-(m,n)} + K\alpha}.
\end{split}
\end{equation}
In different corpus or applications, Equation (\ref{sampler}) can be replaced  with  other Gibbs updating rules or topic models, eg. LFLDA \cite{DBLP:journals/tacl/NguyenBDJ15}. 

\subsection{Topic Embeddings Computing}
Topic embeddings aim to approximate the latent semantic centroids in vector space rather than a multinomial distribution. {\twe} trains topic embeddings after word embeddings have been learned by Skip-Gram. In {\ours}, we use a straightforward way to compute topic embedding for each topic. For the $k$th topic, its topic embedding is computed by averaging all words with their topic assignment $z$ equivalent to $k$, i.e.,
\begin{equation}
\label{topic_cal}
\bm{t}_k = \frac{\sum\limits_{m=1}^M \sum\limits_{n=1}^{N_m} \mathbbm{I}(z_{m,n} = k) \cdot \bm{v}_{w_{m,n}}}{\sum_{w=1}^W n_{k,w}}
\end{equation}
where $\mathbbm{I}(x)$ is indicator function defined as 1 if x is true and 0 otherwise.

Similarly, you can design your own more complex training rule to train topic embedding like TopicVec \cite{DBLP:conf/acl/LiCZM16} and Latent Topic Embedding ({\tt LTE}) \cite{DBLP:conf/coling/JiangSLW16}.
\ignore{ 
This simple way shows the interactions in two aspect.
\begin{itemize} 
\item The parameter $\bm{\varphi}$ make its effect on training topic embedding through $z_{m,n}$ with Gibbs updating rule as in Equation (\ref{sampler}). 
\item Word embeddings is directly used for training topic embeddings so that topic embeddings would be updated along with the words embeddings training step.
\end{itemize}
}

\subsection{Word Embeddings Training}
\label{word_vec}
{\ours} aims to update $\bm{\varphi}$ during word embeddings training. Following the similar optimization as Skip-Gram, hierarchical softmax and negative sampling are used for training the word embeddings approximately due to the computationally expensive cost of the full softmax function which is proportional to vocabulary size $W$. {\ours} uses stochastic gradient descent to optimize the objective function given in Equation (\ref{LTSG_OBJFUNC}). 

The hierarchical softmax uses a binary tree (eg. a Huffman tree) representation of the output layer with the $W$ words as its leaves and, for each node, explicitly represents the relative probabilities of its child nodes. There is a unique path from root to each word $w$ and $node(w, i)$ is the $i$-th node of the path. Let $L(w)$ be the length of this path, then $node(w,1) = root$ and $node(w,L(w)) = w$. Let $child(u)$ be an arbitrary child of node $u$, e.g. left child. By applying hierarchical softmax on $\Pr(w_{m,n+j} \vert T_{w_{m,n}})$ similar to $\Pr(w_{m,n+j} \vert w_{m,n})$ desciebed in Skip-gram \cite{DBLP:conf/nips/MikolovSCCD13}, we can compute the log gradient of $\bm{\varphi}$ as follows,    
\begin{equation}
\label{phi_update}
\begin{split}
&\frac{\partial \log\Pr(w_{m,n+j} \vert T_{w_{m,n}})}{\partial \varphi_{k = z_{m,n}, w = w_{m,n}}} = \frac{1}{L(w_{m,n}-1)} \sum\limits_{i=1}^{L(w_{m,n}) - 1} \\
&\Big[1- h_{i+1}^{w_{m,n+j}}  - \sigma(T_{w_{m,n}} \cdot \bm{v}_i^{w_{m,n+j}})\Big]\bm{t}_k\cdot \bm{v}_i^{w_{m,n+j}},
\end{split}
\end{equation}
where $\sigma(x)=1/(1+\exp(-x))$. Given a path from root to word $w_{m,n+j}$ constructed by Huffman tree, $\bm{v}_i^{w_{m,n+j}}$ is the vector representation of $i$-th node. And $h_{i+1}^{w_{m,n+j}}$ is the Huffman coding on the path defined as $h_{i+1}^{w_{m,n+j}} = \mathbbm{I}\big( node(w_{m,n+j}, i+1) = child(node(w_{m,n+j}, i) \big)$.

Follow this idea, we can compute the gradients for updating the word $w$ and non-leaf node. From Equation (\ref{phi_update}), we can see that $\bm{\varphi}$ is updated by using topic embeddings $v_k$ directly and word embeddings indirectly via the non-leaf nodes in Huffman tree, which is used for training the word embeddings.

\subsection{An overview of our {\ours} algorithm}
\label{CAI}
In this section we provide an overview of our {\ours} algorithm, as shown in Algorithm \ref{alg:LTSG_IILF}. In line 1 in Algorithm \ref{alg:LTSG_IILF}, we run the standard {\lda} with certain iterations and initialize $\bm{\varphi}$ based on Equation (\ref{Phi_cal}). From lines 4 to 6, there are the three steps mentioned in section \ref{LTSG}. From lines 7 to 13, $\bm{\varphi}$ will be updated after training the whole corpus $\bm{\mathcal{D}}$ rather than per word, which is more suitable for multi-thread training. Function $f(\xi,n_{k,w})$ is a dynamic learning rate, defined by $f(\xi,n_{k,w}) = \xi \cdot \log(n_{k,w}) / n_{k,w}$. In line 16, document-topic distribution $\theta_{m,k}$ is computed to model documents.

\renewcommand{\algorithmicrequire}{\textbf{Input:}}
\renewcommand{\algorithmicensure}{\textbf{Output:}}
\begin{algorithm}[h]
	\caption{Latent Topical Skip-Gram on Iterative Interactive Learning Framework}
	\label{alg:LTSG_IILF}
	\begin{algorithmic}[1]
	\Require corpus $\bm{\mathcal{D}}$, \# topics $K$, size of vocabulary $W$, Dirichlet hyperparameters $\alpha, \beta$, \# iterations of {\lda} for initialization $I$, \# iterations of framework IILF $nItrs$, \# Gibbs sampling iterations $nGS$.
	\Ensure $\theta_{m,k}$, $\varphi_{k,w}$, $\bm{v}_w$, $\bm{t}_k$,  $m=1,2,\ldots,M; k=1,2,\ldots,K;w=1,2,\ldots,W$ 
	\State \textbf{Initialization.} Initialize $\varphi_{k,w}$ as in Equation (\ref{Phi_cal}) with $I$ iterations in standard {\lda} as in Equation (\ref{GibbsLDA})
	\State $i \gets 0$
	\While{($i < nItrs$)}
		\State \textbf{Step \ref{Step-1}.} Sample $z_{m,n}$ as in Equation (\ref{sampler}) with $nGS$ iterations
		\State \textbf{Step \ref{Step-2}.} Compute each topic embedding $\bm{t}_k$ as in Equation (\ref{topic_cal})
		\State \textbf{Step \ref{Step-3}.} Train word embeddings with objective function as in Equation (\ref{LTSG_OBJFUNC})
		\State Compute the first-order partial derivatives $\mathcal{L}^\prime(\bm{\mathcal{D}})$
		\State Set the learning rate $\xi$
		\For{($k=1 \to K$)}
			\For{($w=1 \to W$)}
				\State $\varphi_{k,w}^{(i+1)} \gets \varphi_{k,w}^{(i)} + f(\xi,n_{k,w}) \frac{\partial \mathcal{L}^\prime(\bm{\mathcal{D}})}{\partial \varphi_{k,w}}$
			\EndFor
		\EndFor
	\State $i \gets i + 1$
	\EndWhile
	\State Compute each $\theta_{m,k}$ as in Equation (\ref{Theta_cal})	
	\end{algorithmic}
\end{algorithm}

\section{Experiments}
In this section, we evaluate our {\ours} model in three aspects, i.e., contextual word similarity, text classification, and topic coherence.

We use the dataset 20NewsGroup, which consists of about 20,000 documents from 20 different newsgroups. For the baseline, we use the default settings of parameters unless otherwise specified. Similar to {\twe}, we set the number of topic $K=80$ and the dimensionality of both word embeddings and topic embeddings $d = 400$ for all the relative models. In {\ours}, we initialize $\bm{\varphi}$ with $I = 2500$. We perform $nItrs=5$ runs on our framework. We perform $nGS = 200$ Gibbs sampling iterations to update topic assignment with $\alpha = 0.01, \beta = 0.1$.

\subsection{Contextual Word Similarity}
To evaluate contextual word similarity, we use Stanford's Word Contextual Word Similarities (SCWS) dataset introduced by \cite{DBLP:conf/acl/HuangSMN12}, which has been also used for evaluating state-of-art model \cite{DBLP:conf/aaai/LiuLCS15}. There are totally 2,003 word pairs and their sential contexts. For comparison, we conpute the Spearman correlation similarity scores of different models and human judgments.

Following the {\twe} model, we use two scores \verb|AvgSimC| and \verb|MaxSimC| to evaluate the multi-prototype model for contextual word similarity. The topic distribution $\Pr(z|w,c)$ will be infered by regarding $c$ as a document using $\Pr(z|w,c) \propto \Pr(w|z)\Pr(z|c)$. Given a pair of words with their contexts, namely $(w_i,c_i)$ and $(w_j,c_j)$, \verb|AvgSimC| aims to measure the averaged similarity between the two words all over the topics:
\begin{equation}
\label{avgsim}
AvgSimC=\sum\limits_{z,z^\prime \in K} \Pr(z|w_i,c_i)\Pr(z^\prime|w_j,c_j)S(\bm{v}_{w_i}^z,\bm{v}_{w_j}^{z^\prime})
\end{equation}
where $\bm{v}_{w}^z$ is the embedding of word $w$ under its topic $z$ by concatenating word and topic embeddings $\bm{v}_w^z = \bm{v}_w \oplus \bm{t}_z$. $S(\bm{v}_{w_i}^z,\bm{v}_{w_j}^{z^\prime})$ is the cosine similarity between $\bm{v}_{w_i}^z$ and $\bm{v}_{w_j}^{z^\prime}$.

\verb|MaxSimC| selects the corresponding topical word embedding $\bm{v}_w^z$ of the most probable topic $z$ inffered using $w$ in context $c$ as the contextual word embedding, defined as
\begin{equation}
MaxSimc = S(\bm{v}_{w_i}^z,\bm{v}_{w_j}^{z^\prime})
\end{equation}
where 

$z = {\arg\max}_{z}\Pr(z|w_i,c_i)$,  $z^\prime = \arg \max_z \Pr(z|w_j,c_j)$.

We consider the two baselines Skip-Gram and {\twe}. Skip-Gram is a well-known single prototype model and {\twe} is the state-of-the-art multi-prototype model. We use all the default settings in these two model to train the 20NewsGroup corpus.
\begin{table}[!ht]
\caption{Spearman correlation $\rho \times 100$ of contextual word similarity on the SCWS dataset.}\label{SCWS}
\centering
\begin{small}
\begin{tabular}{|c|c|c|}
\hline \bf Model & \multicolumn{2}{c|}{$\rho \times 100$} \\ \hline
Skip-Gram & \multicolumn{2}{c|}{51.1} \\
{\ours}-word & \multicolumn{2}{c|}{52.9} \\
\hline
& \bf \verb|AvgSimC| & \bf \verb|MaxSimC| \\ \hline
 {\twe} & 52.0 & 49.2 \\ 
 {\ours} & 53.5 & 53.0 \\ \hline
\end{tabular}
\end{small}
\end{table}

From Table \ref{SCWS}, we can see that {\ours} achieves better performance compared to the two competitive baseline. It shows that topic model can actually help improving polysemous-word model, including word embeddings and topic embeddings.

\subsection{Text Classification}
\label{task:DC}
In this sub-section, we investigates the effectiveness of {\ours} for document modeling using multi-class text classification. The 20NewsGroup corpus has been divided into training set and test set with ratio 60\% to 40\% for each category.\ignore{We use the dataset 20NewsGroup, which consists of about 20,000 documents from 20 different newsgroups.} We calculate macro-averaging precision, recall and F1-measure to measure the performance of {\ours}.

We learn word and topic embeddings on the training set and then model document embeddings for both training set and testing set. Afterwards, we consider document embeddings as document features and train a linear classifier using Liblinear \cite{DBLP:journals/jmlr/FanCHWL08}. We use $\bm{v}_m$, $\bm{t}_k$, $\bm{v}_w$ to represent document embeddings, topic embeddings, word embeddings, respectively, and model documents on both topic-based and embedding-based methods as shown below.
\begin{itemize}
\setlength{\itemsep}{0.1ex}
\item \textbf{LTSG-theta.} Document-topic distribution $\bm{\theta}_m$ estimated by Equation (\ref{Theta_cal}).
\item \textbf{LTSG-topic.} $\bm{v}_m = \sum_{k=1}^K \theta_{m,k} \cdot \bm{t}_k$.
\item \textbf{LTSG-word.} $\bm{v}_m = (1/N_m) \sum_{n=1}^{N_m} \bm{v}_{w_{m,n}}$.
\item \textbf{LTSG.} $\bm{v}_m =(1/N_m) \sum_{n=1}^{N_m} \bm{v}_{w_{m,n}}^{z_{m,n}}$, where contextual word is simply constructed by $\bm{v}_{w_{m,n}}^{z_{m,n}} = \bm{v}_{w_{m,n}} \oplus \bm{t}_{z_{m,n}}$.  
\end{itemize}

We consider the following baselines, bag-of-word (BOW) model, {\lda}, Skip-Gram and {\twe}. The BOW model represents each document as a bag of words and use TFIDF as the weighting measure. For the TFIDF model, we select top 50,000 words as features according to TFIDF score. {\lda} represents each document as its inferred topic distribution. In Skip-Gram, we build the embedding vector of a document by simply averaging over all word embeddings in the document. The experimental results are shown in Table \ref{text-classification}.
\begin{table}[!ht]
\caption{\label{text-classification} Evaluation results of multi-class text classification.}
\centering
\begin{small}
\begin{tabular}{|c|c|c|c|c|}
\hline \bf Model & \bf Accuracy & \bf Precision & \bf Recall & \bf F1-measure \\ \hline
BOW 			& 79.7 & 79.5 & 79.0 & 79.2 \\
{\lda} 			& 72.2 & 70.8 & 70.7 & 70.7 \\
Skip-Gram 	& 75.4 & 75.1 & 74.7 & 74.9 \\
{\twe} 		& 81.5 & 81.2 & 80.6 & 80.9 \\
\hline
LTSG-theta 	& 72.6 & 71.9 & 71.2 & 70.2 \\
LTSG-topic 	& 73.8 & 73.0 & 72.4 & 71.3 \\
LTSG-word  	& 81.2 & 80.6 & 80.2 & 80.2 \\
LTSG & \textbf{82.8} & \textbf{82.4} & \textbf{81.8} & \textbf{81.8} \\
\hline
\end{tabular}
\end{small}
\end{table}

From Table \ref{text-classification}, we can see that, for topic modeling, LTSG-theta and LTSG-topic perform better than {\lda} slightly. For word embeddings, LTSG-word significantly outperforms Skip-Gram. For topic embeddings using for multi-prototype word embeddings, LTSG also outperforms state-of-the-art baseline {\twe}. This verifies that topic modeling, word embeddings and topic embeddings can indeed impact each other in {\ours}, which lead to the best result over all the other baselines. 
 
\subsection{Topic Coherence}
\label{task:TC}
\begin{table*}[tb]
\caption{\label{topic_coherence} Top words of some topics from {\ours} and {\lda} on 20NewsGroup for $K = 80$.}
\centering
\begin{tabular}{|cc|cc|cc|cc|}
\hline {\ours} & {\lda} & {\ours} & {\lda} & {\ours} & {\lda} & {\ours} & {\lda} \\ \hline
image & image  & jet & printer & stimulation & doctor & anonymous & list \\
jpeg & files & ink & good & diseases & disease & faq & mail \\
gif & color & laser & print & disease & coupons & send & information \\
format & gif & printers & font & toxin & treatment & ftp & internet \\
files & jpeg  & deskjet & graeme & icts & pain & mailing & send \\
file & file & ssa & laser & newsletter & medical & server & posting \\
convert & format & printer & type & staffed & day & mail & email \\
color & bit & noticeable & quality & volume & microorganisms & alt & group \\
formats & images & canon & printers & health & medicine & archive & news \\
images & quality & output & deskjet & aids & body & email & nonymous \\
\hline
-75.66 & -88.76 & -91.53 & -119.28 & -66.91 & -100.39 & -78.23 & -95.47 \\
\hline
\end{tabular}
\end{table*}
In this section, we evaluate the topics generated by {\ours} on both quantitative and qualitative analysis. Here we follow the same corpus and parameters setting in section \ref{task:DC} for LSTG model.
\paragraph*{Quantitative Analysis} Although perplexity (held-out likehood) has been widely used to evaluate topic models, \cite{chang2009reading} found that perplexity can be hardly to reflect the semantic coherence of individual topics. Topic Coherence metric \cite{DBLP:conf/emnlp/MimnoWTLM11} was found to produce higher correlation with human judegments in assessing topic quality, which has become popular to evaluate topic models \cite{DBLP:conf/icml/AroraGHMMSWZ13,DBLP:conf/icml/Chen014}. A higher topic coherence score indicates a more coherent topic.

We compute the score of the top 10 words for each topic. We present the score for some of topics in the last line of Table \ref{topic_coherence}. By averaging the score of the total 80 topics, {\ours} gets -92.23 compared with -108.72 of {\lda}. We can conclude that {\ours} performs better than {\lda} in finding higher quality topics.
\paragraph*{Qualitative Analysis} Table \ref{topic_coherence} shows top 10 words of topics from {\ours} and {\lda} model on 20NewsGroup. The words in this two models are ranked based on the probability distribution $\bm{\varphi}$ for each topic. As shown, {\ours} is able to capture more concrete topics compared with general topics in {\lda}. For the topic about ``image'', {\ours} shows about image convertion on different format, while {\lda} shows the image quality of different format. In topic ``printer'', {\ours} emphasizes the different technique of printer in detail and {\lda} generally focus on ``good quality'' of printing. In topic about ``mail'', {\ours} gives a way to build a mail server while {\lda} generally concerns about attribute of email.

\section{Releated Work}
Rencently, researches on cooperating topic models and vector representations have made great advances in NLP community. \cite{DBLP:conf/naacl/XieYX15} proposed a Markov Random Field regularized {\lda} model (\mlda) to incorporate word similarities into topic modeling. The {\mlda} model encourages similar words to share the same topic so as to learn more coherent topics. \cite{DBLP:conf/acl/DasZD15} proposed Gaussian {\lda} to use pre-trained word embeddings in Gibbs sampler based on multivariate Gaussian distributions. \cite{DBLP:journals/tacl/NguyenBDJ15} proposed LFLDA which is modeled as a mixture of the conventional categorical dirtribution and an embedding link function. These works have given the faith that vector representations are capable of helping improving topic models. On the contrary, vector representations, especially topic embeddings, have been promoted for modeling documents or polysemy with great help of topic models. For examples, \cite{DBLP:conf/aaai/LiuLCS15} used topic model to globally cluster the words into different topics accroding to their context for learning better multi-prototype word embeddings. \cite{DBLP:conf/acl/LiCZM16} proposed generative topic embedding (TopicVec) model that replaces categorical distribution in LDA with embedding link function. However, these models do not show close interactions among topic models, word embeddings and topic embeddings. Besides, these researches lack of investigation on the influence of topic model on word embeddings.

\section{Conclusion and Future Work}
In this paper, we introduce a general framework to make topic models and vector representations mutually help each other. We propose a basic model Latent Topical Skip-Gram (\ours) which shows that {\lda} and Skip-Gram can mutually help improve performance on different task. The experimental results show that {\ours} achieves the competitive results compaired with the state-of-art models. Especially, we can make a conclusion that topic model helps promoting word embeddings in {\ours} model.

We consider the following future research directions:
\begin{inparaenum}[I)]
\item The number of topics must be pre-defined and Gibbs sampling is time-consuming for training large-scale data with using single thread. we will investigate non-parametric topic models \cite{doi:10.1198/016214506000000302} and parallel topic models \cite{DBLP:journals/tist/LiuZCS11}. 
\item There are many topic models and word embeddings models have been proposed to use in various tasks and specific domains. We will construct a package which can be convenient to extend with other models to our framework by using the interfaces. 
\item {\ours} could not deal with unseen words in new documents, we may explore techniques to train word embeddings and topic assigments for the unseen words like Gaussian {\lda} \cite{DBLP:conf/acl/DasZD15}.
\item We wish to evaluate topic embeddings directly similar to topic coherence task.
\end{inparaenum}

\bibliographystyle{named}
\bibliography{ltsg}

\end{document}